\documentclass{article}
\usepackage{spconf,amsmath,graphicx}
\usepackage{graphicx,subfigure,subfig,epsfig,amsfonts,amsmath,amssymb,lscape,color,url}
\usepackage[noadjust]{cite}

\title{Tree-Structure Bayesian Compressive Sensing for Video}
\name{Xin Yuan\thanks{xin.yuan@duke.edu}, Patrick Llull, David J. Brady, and Lawrence Carin}
\address{Department of Electrical and Computer Engineering, Duke University,
Durham, NC, 27708, USA}
\begin{document}
%
\maketitle
\begin{abstract}
A Bayesian compressive sensing framework is developed for video reconstruction based on the color coded aperture compressive temporal imaging (CACTI) system.
By exploiting the three dimension (3D) tree structure of the wavelet and  Discrete Cosine Transformation (DCT) coefficients, a Bayesian compressive sensing inversion algorithm is derived to reconstruct (up to 22) color video frames from a {\em single} monochromatic compressive measurement.
Both simulated and real datasets are adopted to verify the performance of the proposed algorithm.
\end{abstract}
\section{Introduction}
\label{sec:intro}

The mathematical theory of compressive sensing (CS) \cite{cs_Donoho06} asserts that
one can acquire signals from measurements whose rate is much
lower than the total bandwidth. Whereas the CS theory is now
well developed, challenges concerning hardware implementations \cite{Sankaranarayanan10ECCV,Sankaranarayanan12ICCP,Wakin06PCS,Duarte08SPM,Kittle10AO,Veeraraghavan11TPAMI}
of CS-based acquisition devices, especially in optics, have only
started being addressed.
This paper will introduce a color video CS camera capable of 
capturing low-frame-rate measurements at acquisition, with high-frame-rate video recovered subsequently via computation (decompression of the measured data).


The Coded Aperture Compressive Temporal Imaging (CACTI) \cite{Patrick13OE} system uses a moving binary mask pattern to modulate a video sequence within the integration time $\Delta_t$ many times prior to integration by the detector.  The number of high-speed frames recovered from a coded-exposure measurement depends on the speed of video modulation.
Within the CACTI framework, modulating the video $n_t$ times per second corresponds to moving the mask $n_t$ pixels within the integration time $\Delta_t$.
If $n_{t}$ frames are to be recovered per compressive measurement by a camera collecting data at $1/\Delta_{t}$ frames-per-second (fps), the time variation of the code is required to be $n_{t}/\Delta_{t}$ fps. The liquid-crystal-on-silicon (LCoS) modulator used in \cite{Hitomi11ICCV,Reddy11CVPR} can modulate as fast as $3000$ fps by pre-storing the exposure codes, but, because the coding pattern is continuously changed at each pixel throughout the exposure, it requires considerable energy consumption ($>3W$). The mechanical modulator in \cite{Patrick13OE}, by contrast, modulates the exposure through periodic mechanical translation of a single mask (coded aperture), using a pizeoelectronic translator that consumes minimal energy ($\sim 0.2W$).
The coded aperture compressive temporal imaging (CACTI) \cite{Patrick13OE} now has been extended to the color video~\cite{Yuan14CVPR}, which can capture ``R", ``G" and ``B" channels of the context. By appropriate reconstruction algorithms \cite{Yang14GMM,Yuan13ICIP,Liao12GAP}, we can get $n_t$ frames color video from a single gray-scale measurement.


\begin{figure}[htbp!]
\begin{center}
   \includegraphics[width=1.0\linewidth,height = 4cm]{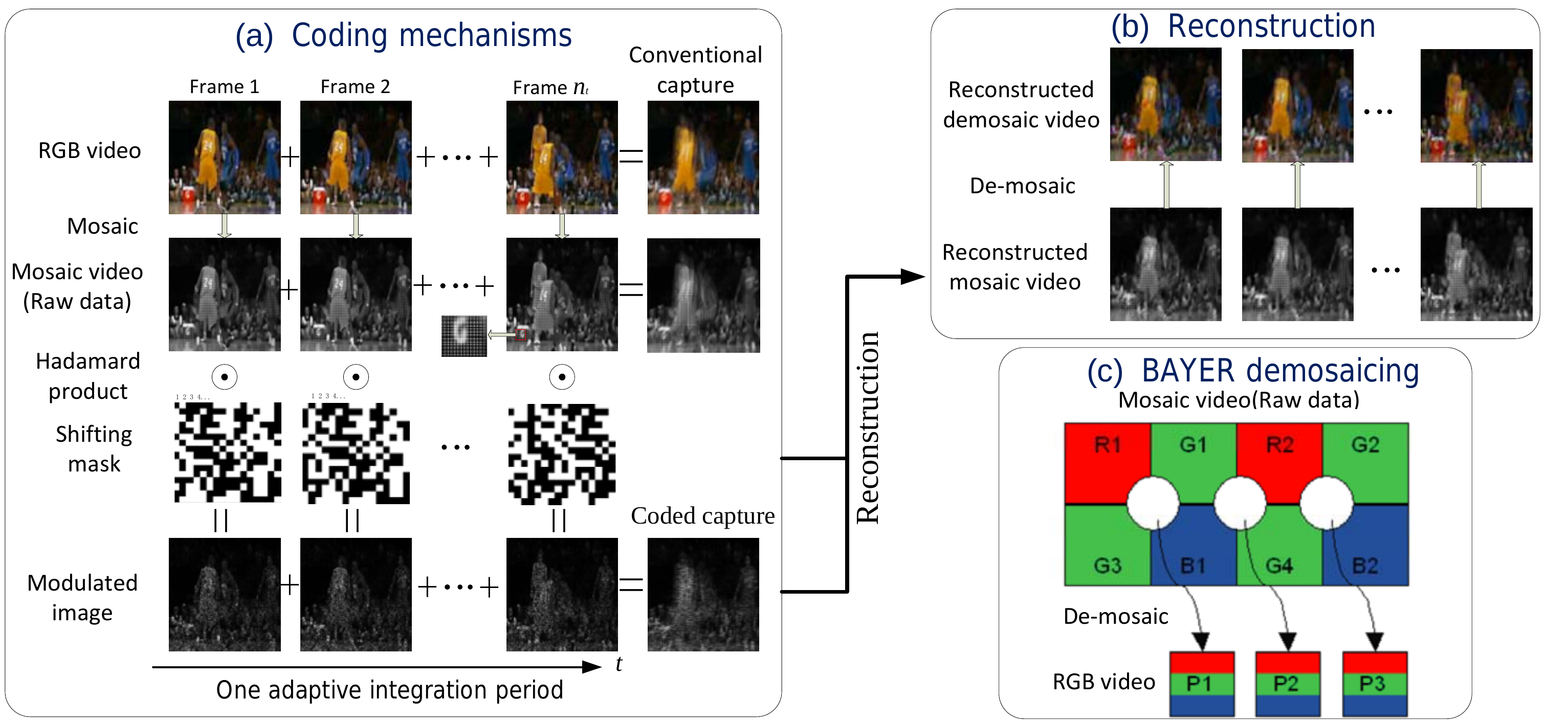}
\end{center}
\vspace{-2mm}
\caption{\small{
(a) First row shows $n_{t}$ color (RGB) frames of the original high-speed video;
second shows each color frame rearranged into a Bayer-filter mosaic;
third row depicts the (horizontally) moving mask used to modulate the high-speed frames (black is zero, white is one); fast translation manifested by a pizeoelectronic translator.
Fourth row shows the modulated frames, whose sum gives a single coded-exposure photo.
(b) Recovered RGB frames arranged into a Bayer-filter mosaic (second row), which is de-mosaicked to give the color frames (first row). (c) The de-mosaicing process \cite{Guppy2012}.}}
\label{Fig:dec}
\vspace{-3mm}
\end{figure}

While numerous algorithms have been used for CS inversion, the Bayesian CS algorithm \cite{Ji08SPT} has been shown with significant advantages of providing a full posterior distribution.
This paper develops a new Bayesian inversion algorithm to reconstruct videos based on raw measurements acquired by the color-CACTI camera. By exploiting the hybrid three dimensional (3D) tree-structure of the wavelet and DCT (Discrete Cosine Transform) coefficients, we have developed a Hidden Markov tree (HMT) \cite{Crouse98SPT} model in the context of a Bayesian framework. 
Research in \cite{Baraniuk10ITT,He09SPT,He10SPL,Som12TSP,Song13TSP} has shown that by employing the  HMT structure of an image, the CS measurements can be reduced.
This paper extends this HMT to 3D and a sophisticated 3D tree-structure is 
developed for video CS, with color-CACTI shown as an example.
Experimental results with both simulated and real datasets verify the performance of the proposed algorithm.
The basic model and inversion method may be applied to any of the compressive video cameras discussed above.



\vspace{-3mm}
\section{Color-CACTI}
\vspace{-3mm}
\subsection{Coding Scenario}
\vspace{-2mm}
Let $z(x,y,t)$ be the continuous/analog spatiotemporal volume of the video being measured; $\phi(x-r(t),y-s(t))$ represents a moving mask (code) with $(r(t),s(t))$ denoting its spatial translation at time $t$; and $h(x,y)$ denotes the camera spatial sampling function, with spatial resolution $\Delta_{x}\Delta_{y}$. The coded aperture compressive camera system modulates each temporal segment of duration $\Delta_{t}$ with the moving mask (the motion is periodic with the period equal to $\Delta_{t}$), and collapses (sums) the coded video into a single photograph ($\forall$ $l\geq1$):
\begin{eqnarray}\label{eq:CACTI-measurement}
y_{ijl}&=& \int_{0}^{\Delta_{t}}dt\int_{0}^{n_{x}\Delta_{x}}dx\int_{0}^{n_{y}\Delta_{y}}dy~ z(x,y,t\!+\!(l\!-\!1)\Delta_{t})\nonumber \\
&&\times \phi(x\!-\!r(t),y\!-\!s(t))\,h\left(x\!-\!i\Delta_{x},y\!-\!j\Delta_{y}\right),
\end{eqnarray}
$i=1,\dots,n_{x}$ and $j=1,\dots,n_{y}$, with the detector size $n_x\times n_y$ pixels. The set of data $\{y_{ijl}\}$, which below we represent as $\mathbf{Y}_{l}$, corresponds to the $l$th compressive measurement. The code/mask $\phi(x,y)$ is here binary, corresponding to photon transmission and blocking (see Figure \ref{Fig:dec}).


\vspace{-3mm}
\subsection{Measurement Model}
\vspace{-2mm}
Denote $z_{ijkl}=z\left(i\Delta_{x},j\Delta_{y},\frac{k\Delta_{t}}{n_{t}}\!+\!(l\!-\!1)\Delta_{t}\right)$, defining the original continuous video $z(x,y,t)$ sampled in space $(i,j)$ and in time ($n_t$ discrete temporal frames, $k=1,\dots, n_t$, within the time window of the $l$th compressive measurement). We also define 
\begin{eqnarray}\label{eq:CACTI-sensing-matrix}
\phi_{ijk}&=&\int_{0}^{n_{x}\Delta_{x}}\!\!\!\!\int_{0}^{n_{y}\Delta_{y}}\!\!\!\!\phi(x\!-\!r(t),y\!-\!s(t))\left|_{t=\frac{k\Delta_{t}}{n_{t}}}\right.\nonumber\\
&&\times h\left(x\!-\!i\Delta_{x},y\!-\!j\Delta_{y}\right)dxdy.
\end{eqnarray}
We can rewrite (\ref{eq:CACTI-measurement}) as
\vspace{-3mm}
\begin{eqnarray}\label{eq:CACTI-measurement-discrete}
y_{ijl}&=&\sum_{k=1}^{n_{t}}z_{ijkl}\,\phi_{ijk}+e_{ijl},\,\,\forall\,i,j,\\
\mathbf{Y}_{l}&=&\sum_{k=1}^{n_{t}}\boldsymbol\Phi_{k}\odot\mathbf{Z}_{kl}+\mathbf{E}_{l},
\vspace{-3mm}
\end{eqnarray}
where $e_{ijl}$ is an added noise term, $\mathbf{Y}_{l},\boldsymbol\Phi_{k},\mathbf{Z}_{kl},\mathbf{E}_{l}\in\mathbb{R}^{n_{x}\times{}n_{y}}$, and $\odot$ denotes element-wise multiplication (Hadamard product). In (\ref{eq:CACTI-measurement-discrete}), $\mathbf{\Phi}_k$ denotes the mask/code at the $k$th shift position (approximately discretized in time), and $\mathbf{Z}_{kl}$ is the underlying video, for video frame $k$ within CS measurement $l$. Dropping subscript $l$ for simplicity, (\ref{eq:CACTI-measurement-discrete}) can be written as
\vspace{-3mm}
\begin{eqnarray}\label{eq:CACTI-measurement-discrete-vec}
\mathbf{y}&=& \mathbf{H}\mathbf{x} +\mathbf{e},\\
\mathbf{H}&=&\left[\mathrm{diag}(\mathrm{vec}(\boldsymbol\Phi_{1})),\cdots,\mathrm{diag}(\mathrm{vec}(\boldsymbol\Phi_{n_{t}}))\right]\\
\mathbf{x}&=&\mathrm{vec}([\mathbf{Z}_{1},\cdots,\mathbf{Z}_{n_{t}}]),
\vspace{-3mm}
\end{eqnarray}
where $\mathbf{y}=\mathrm{vec}(\mathbf{Y})$ and $\mathrm{vec}(\cdot)$ is standard vectorization.

\vspace{-3mm}
\subsection{Mosaicing and De-mosaicing of Color Video}
\vspace{-2mm}
We record temporally compressed measurements for RGB colors on a Bayer-filter mosaic, where the three colors are arranged in the pattern shown in the right bottom of Figure \ref{Fig:dec}. The single coded image is partitioned into four components, one for R and B and two for G (each is $1/4$ the size of the original spatial image). The CS recovery (video from a single measurement) is performed separately on these four mosaiced components, prior to demosaicing as shown in Figure \ref{Fig:dec}(b). One may also jointly perform CS inversion on all 4 components, with the hope of sharing information on the importance of (here wavelet and DCT) components; this was also done, and the results were very similar to processing R, B, G1 and G2 separately.
Note that this is the key difference between color-CACTI and the previous work of CACTI in \cite{Patrick13OE}.

\vspace{-3mm}
\section{Bayesian Compressive Sensing for Video Reconstruction}
\vspace{-3mm}
\subsection{3D Tree Structure of Wavelet Coefficients}
\vspace{-2mm}
An image's zero-tree structure \cite{Shapiro93SPT} has been investigated thoroughly  since the advent of wavelets \cite{Mallat}.
The 3D wavelet tree structure of video, an extension of the 2D image, has also attracted extensive attention in the literature \cite{Chen96SPIE}.
Xiong {\em et al.} introduced a tree-based representation to characterize the block-DCT transform associated
with JPEG \cite{Xiong96SPL}.
For the video representation, we here use the wavelet in space and DCT in time.

\begin{figure}[h!]
\begin{center}
   \includegraphics[scale = 0.25]{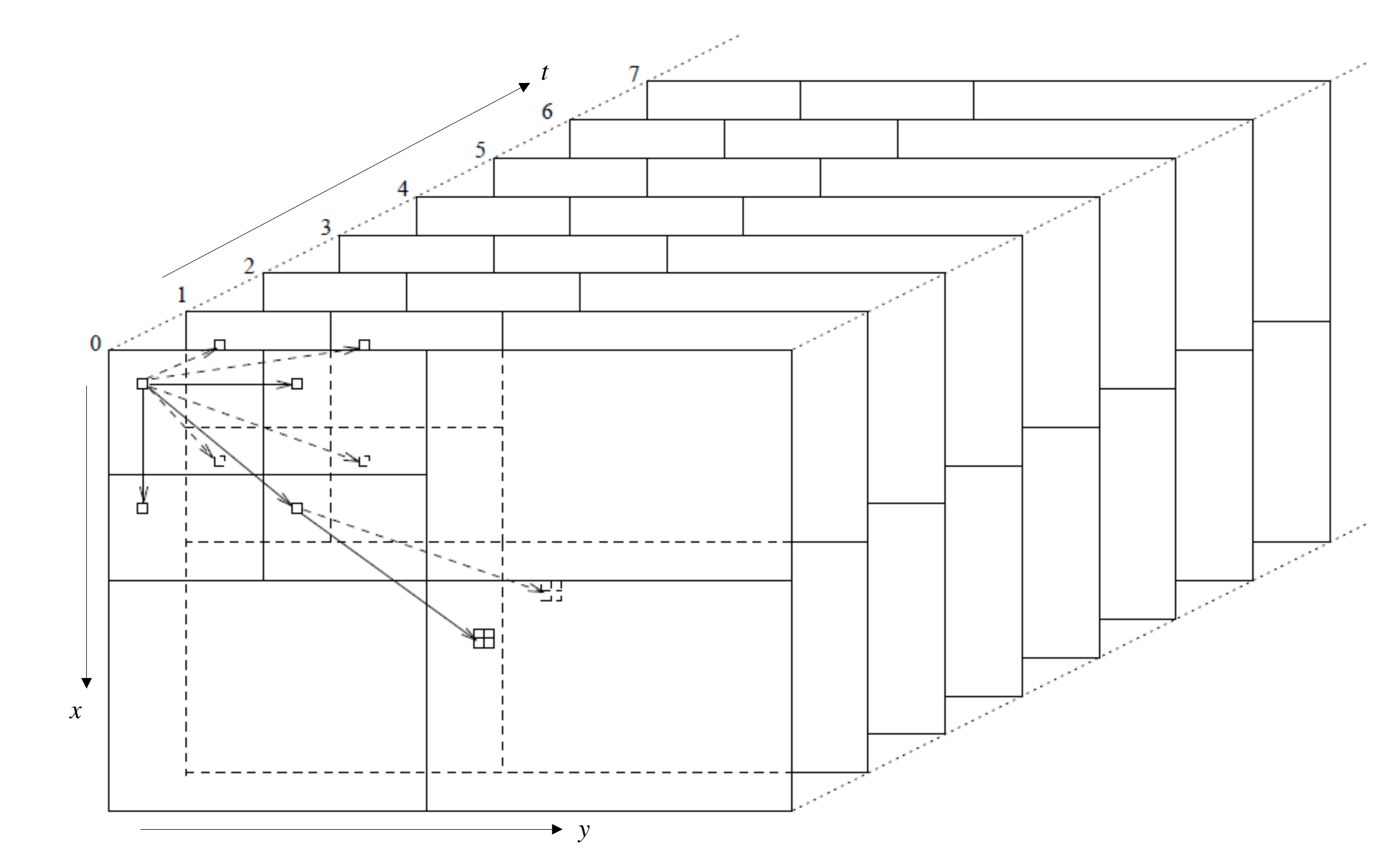}
\end{center}
\vspace{-2mm}
\caption{\small{3D tree structure of wavelet coefficients.}}
\label{Fig:3d_tree}
\vspace{-2mm}
\end{figure}

Considering the video sequence has $n_t$ frames with spatial $n_xn_y$ pixels, 
and let $(i,j, t)$ denote the indices of the DCT/Wavelet coefficients.
Assume there are $L$ levels (scales) of the coefficients ($L=3$ in figure \ref{Fig:3d_tree}).
The parent-children linkage of the coefficients are as follows:
a) a root-node $(i,j,t)$ has 7 children, $(i+b_x,j,t),(i,j+b_y,t),(i,j,t+b_t),(i+b_x,j+b_y,t),(i+b_x,j,t+b_t),(i,j+b_y,t+b_t),(i+b_x,j+b_y, t+b_t)$, where $(b_x,b_y,b_t)$ denotes the size of scaling (LL) coefficients;
b) an internal node $(i,j,t)$ has 8 children
$(2i,2j,2t),(2i+1,2j,2t),(2i,2j+1,2t),(2i,2j,2t+1),(2i+1,2j+1,2t),(2i+1,2j,2t+1),(2i,2j+1,2t+1),(2i+1,2j+1,2t+1)$;
and c) a leaf-node has no children.

When the tree structure is used in 3D DCT, we consider the block size of the 3D DCT is $\{P_x, P_y, P_t\}$, and $P_x = P_y = P_t = 2^L $. The parent-children linkage is the same as with the wavelet coefficients \cite{Xiong96SPL}.

The properties of wavelet coefficients that lead to the Bayesian model derived in the following section are \cite{Crouse98SPT}:\\
1) Large/small values of wavelet coefficients generally persist across the scales of the wavelet tree (the two states of the binary part of the model developed in the following section). \\
2) Persistence becomes stronger at finer scales (the confidence of the probability of the binary part is proportional to the number of coefficients at that scale). \\ 
3) The magnitude of the wavelet coefficients {\em decreases exponentially} as we move to the finer scales.
In this paper, we  use a multiplicative gamma prior \cite{Bhattacharya11MuGam}, a typical shrinkage prior, for the non-zero wavelet coefficients at different scale to embed this decay.

\subsection{Statistical Bayesian Model}
Let $\mathbf{F}_{x}\in\mathbb{R}^{n_{x}\times{}n_{x}}$, $\mathbf{F}_{y}\in\mathbb{R}^{n_{y}\times{}n_{y}}$,  $\mathbf{F}_{t}\in\mathbb{R}^{n_{t}\times{}n_{t}}$  be orthonormal matrices defining bases such as wavelets or the DCT \cite{Mallat}. Define
\begin{eqnarray}
\boldsymbol{\theta} & =& \left(\mathbf{F}_{t}^{T}\otimes\mathbf{F}_{y}^{T}\otimes\mathbf{F}_{x}^{T}\right){\bf x}, \\
\boldsymbol{\Psi}  &=& {\bf H}\left(\mathbf{F}_{t}\otimes\mathbf{F}_{y}\otimes\mathbf{F}_{x}\right),
\end{eqnarray}
where  $\boldsymbol{\theta}$ symbolizes the 3D wavelet/DCT coefficients corresponding to ${\bf F}_x, {\bf F}_y$ and ${\bf F}_t$ and $\otimes$ denotes the Kronecker product.
It is worth noting here the $\boldsymbol{\Psi}$ is the 3D transform of the projection matrix ${\bf H}$. 
Unlike the model used in \cite{He09SPT,He10SPL,Baraniuk10ITT}, where the projection matrix is put directly on the wavelet/DCT coefficients,
in the coding strategy of color-CACTI, we get the projection matrix  ${\boldsymbol \Phi}$ from the hardware by capturing the response of the mask at different positions. Following this, we transform ${\bf H}$
row-by-row to the wavelet/DCT domain, to obtain $\boldsymbol{\Psi}$.

The measurement noise is modeled as zero mean Gaussian with precision matrix (inverse of the covariance matrix) $\alpha_0 {\bf I}$, where ${\bf I}$ is the identity matrix. 
We have:
\begin{eqnarray}
{\bf y}|{\boldsymbol{\theta}},\alpha_0 &\sim& {\cal N}({\boldsymbol{\Psi}} {\boldsymbol{\theta}},\alpha_0^{-1}{\bf I})  
\end{eqnarray}
To model the sparsity of the 3D coefficients of wavelet/DCT, the {\em spike-and-slab} prior is imposed on ${\boldsymbol{\theta}}$ as:
\begin{eqnarray}
{\boldsymbol{\theta}} &=& {\bf z} \odot {\bf w},
\end{eqnarray}
where ${\bf w}\in {\mathbb R}^{n_xn_yn_t}$ is a vector of non-sparse coefficients and ${\bf z}$ is a binary vector (zero/one indicators) denoting the two state of the HMT \cite{Crouse98SPT}, with ``zero" signifying the ``low-state" in the HMT and ``one" symbolizing the ``high-state".
Note when the coefficients lie in the ``low-state", they are explicitly set to zero, which leads to the sparsity.

To model the linkage of the tree structure across the scales of the wavelet/DCT, we use the the binary vector, $\bf z$, which is drawn from a Bernoulli distribution.
The parent-children linkage is manifested by the probability of this vector.
We model ${\bf w}$ is drawn from a Gaussian distribution with the precision modeled as a multiplicative Gamma prior.

The full Bayesian model is:
\begin{eqnarray}
{\bf y}|{\boldsymbol{\theta}},\alpha_0 &\sim& {\cal N}({\boldsymbol{\Psi}} {\boldsymbol{\theta}},\alpha_0^{-1}{\bf I})  \\
{\boldsymbol{\theta}} &=& {\bf z} \odot {\bf w},  \\
w_{i,\ell} &\sim& {\cal N}(0, \alpha^{-1}_{\ell}), \\
z_{i,\ell} &\sim& {\rm Bernoulli}(\pi_{i,\ell}), \\
 \pi_{i,\ell}&=&\left\{\begin{array}{rl}
\pi^{\ell}, & \text {if } \ell=0,1 \\
\pi^{p0}, & \text {if } 2\le \ell\le L, z_{pa(i,\ell)} = 0,\\
\pi^{p1}, & \text {if } 2\le \ell\le L, z_{pa(i,\ell)} = 1,
\end{array}\right. 
\end{eqnarray}
where $\{i,\ell\}$ denotes the $i$th component at level $\ell$,
and $\ell=0$ denotes the scaling coefficients of wavelet (or DC level of a DCT).
\begin{eqnarray}
\alpha_0 &\sim& {\rm Gamma}(a_0,b_0) \\
\alpha_\ell &=& \prod_{j=0}^{\ell}{\tau}_j, \\
\tau_{\ell} &\sim& {\rm Gamma}(c_0,d_0), ~~ \ell=0,\dots, L \\
\pi^{\ell} &\sim& {\rm Beta}(e^{\ell}, f^{\ell}), ~~~~ \ell=0,1\\
\pi^{p0} &\sim& {\rm Beta}(e^{p0}, f^{p0}), ~~~~ 2\le \ell \le L \\
\pi^{p1} &\sim& {\rm Beta}(e^{p1}, f^{p1}), ~~~~ 2\le \ell \le L 
\end{eqnarray}

In the experiments, we use the following settings:
\begin{eqnarray}
e^0 = 1,&& f^0 = 0;  \\
e^1 = 0.9N_{\ell},&& f^1 = 0.1N_{\ell}; \\
e^{p0} = \frac{1}{N}N_{\ell}, && f^{p0} = \frac{N-1}{N}N_{\ell}; \\
e^{p1} = 0.5N_{\ell}, && f^{p1} = 0.5N_{\ell},
\end{eqnarray}
where $N_{\ell}$ is the number of coefficients at $\ell$th level, and $N$ is the length of $\boldsymbol{\theta}$.
\vspace{-3mm}
\subsection{INFERENCES}
\vspace{-2mm}
We developed the variational Bayesian methods to infer the parameters in the model as in \cite{He10SPL}. The posterior inference of $\alpha_{\ell}$, thus $\tau_{\ell}$ is different from the model in \cite{He10SPL}, and we show it below:
\vspace{-3mm}
\begin{eqnarray}
\langle\tau_{\ell}\rangle &=& \frac{c_0 + 0.5\sum_{j=0}^{\ell} N_{\ell}}{d_0 + 0.5\sum_{j=0}^{\ell} \sum_{i=1}^{N_{j}}\langle w^2_{j,i}\rangle},
\vspace{-3mm}
\end{eqnarray}
where $\langle \cdot \rangle$ denotes the expectation in $\langle \hspace{0.02in} \rangle$.

\section{Experimental Results}
Both simulated and real datasets are adopted to verify the performance of the proposed model for video reconstruction.
The hyperparameters are setting as $a_0=b_0=c_0=d_0=1e-6$; the same used in \cite{He09SPT,He10SPL}.
Best results are found when ${\bf F}_x$ and ${\bf F}_y$ are wavelets (here the Daubechies-8 \cite{Mallat}) and ${\bf F}_t$ corresponds to a DCT.
The proposed tree-structure Bayesian CS inversion algorithm is compared with the following algorithms:
$i$) Generalized alternating projection (GAP) algorithm \cite{Liao12GAP,Yuan13ICIP};
$ii$) two-step iterative shrinkage/thresholding (TwIST) \cite{Bioucas-Dias2007TwIST} (with total variation norm); 
$iii$) K-SVD \cite{Aharon06TSP} with orthogonal matching pursuit (OMP) \cite{Tropp04ITT} used for inversion; 
$iv$) a Gaussian mixture model (GMM) based inversion algorithm \cite{Yang14GMM}; and 
$v$) the linearized Bregman algorithm \cite{Yin08bregman}.
The $\ell_1$-norm of DCT or wavelet coefficients is adopted in linearized Bregman and GAP with the same transformation as the proposed model.
GMM and K-SVD are patch-based algorithms and we used a separate dataset for training purpose.
A batch of training videos were used to pre-train K-SVD and GMM, and we selected the best reconstruction results for presentation here.

\vspace{-3mm}
\subsection{Simulation Datasets}
\vspace{-1mm}
We consider a scene in which a basketball player performs a dunk; this video is challenging due to the complicated motion of the basketball players and the varying lighting conditions; see the example video frames in Figure \ref{Fig:dec}(a). 
We consider a binary mask, with 1/0 coding drawn at random Bernoulli(0.5); the code is shifted spatially via the coding mechanism in Figure \ref{Fig:dec}(a)), as in our physical camera. 
The video frames are $256\times 256$ spatially, and we choose $n_t=8$.
It can be seen clearly that the proposed tree-structure Bayesian CS algorithm demonstrates improved PSNR performance for the inversion.

\begin{figure}[h!]
\vspace{-3mm}
\begin{center}
   \includegraphics[scale =0.24]{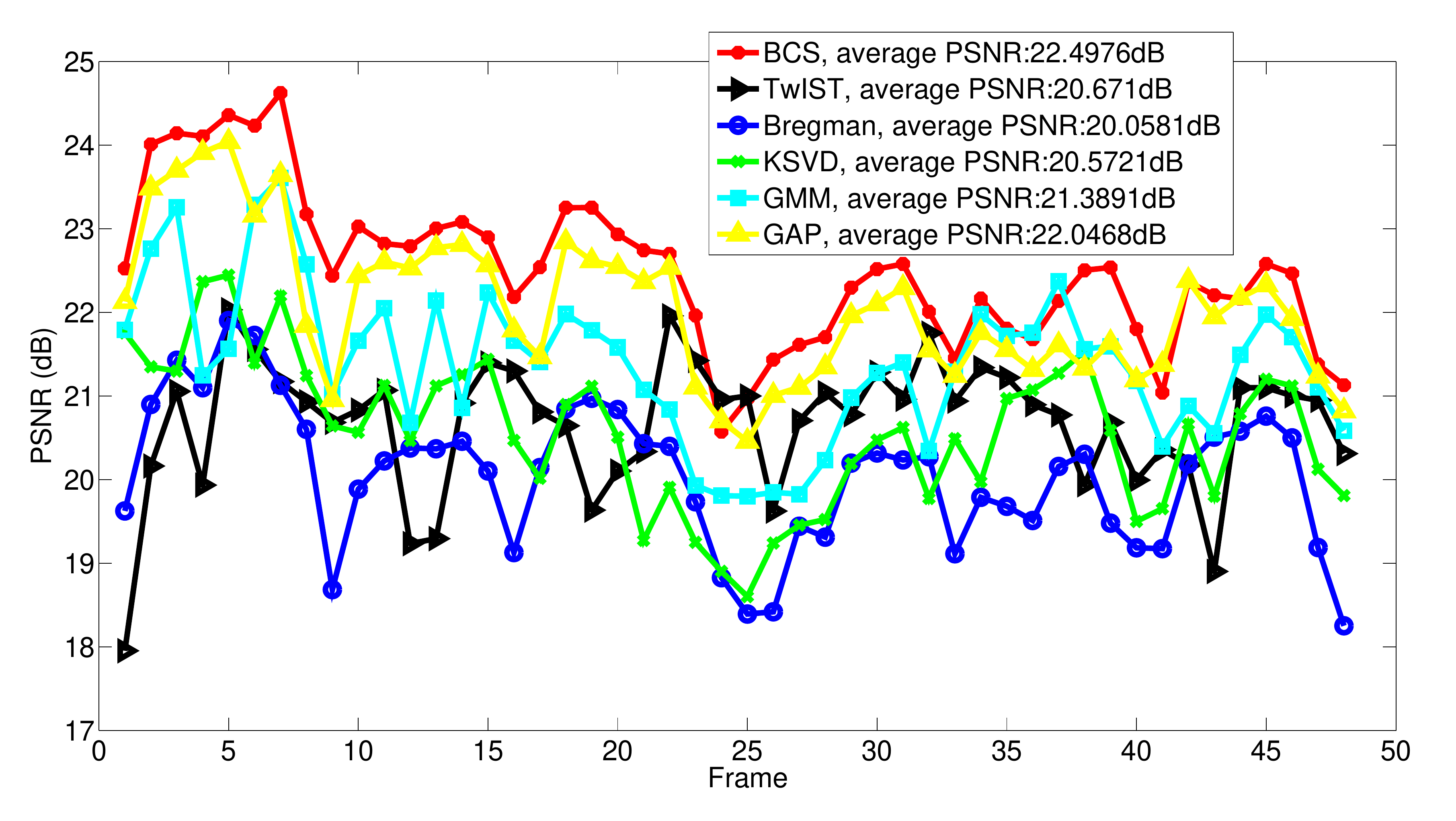}
\end{center}
\vspace{-6mm}
\caption{\small{PSNR comparison of proposed tree-structure Bayesian CS inversion method, GAP, TwIST, linearized Bregman, K-SVD, and GMM algorithms with the simulated dataset.}}
\label{Fig:dunk}
\vspace{-5mm}
\end{figure}

\vspace{-3mm}
\subsection{Real Datasets}
\begin{figure}[h!]
\vspace{-3mm}
\begin{center}
   \includegraphics[scale =0.15]{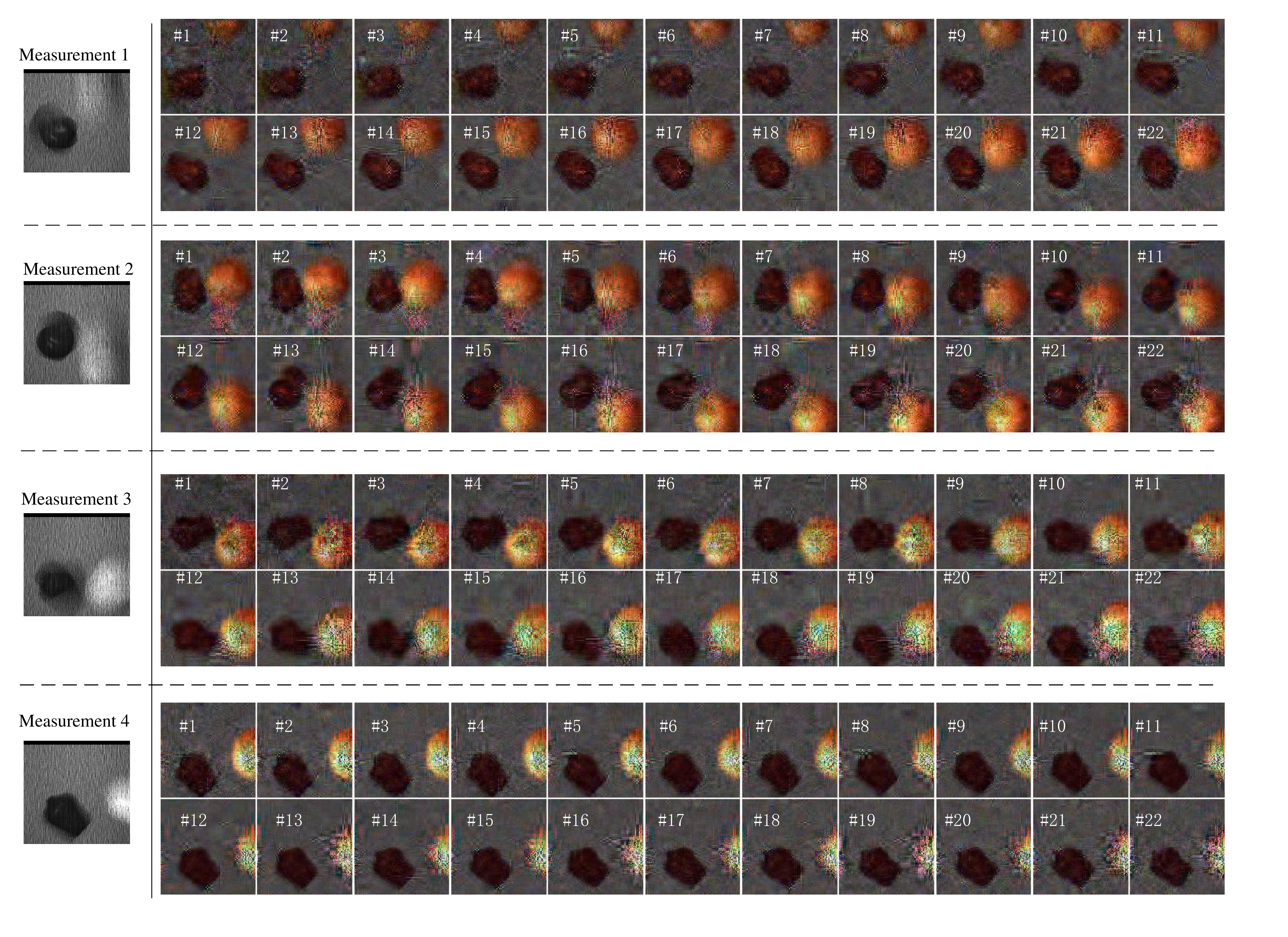}
\end{center}
\vspace{-5mm}
\caption{\small{Reconstruction results of real dataset with ``plastic fruits" (a yellow orange and a red apple) falling and rebounding.}}
\label{Fig:3Balls}
\vspace{-4mm}
\end{figure}

\begin{figure}[h!]
\vspace{-3mm}
\begin{center}
   \includegraphics[scale =0.2]{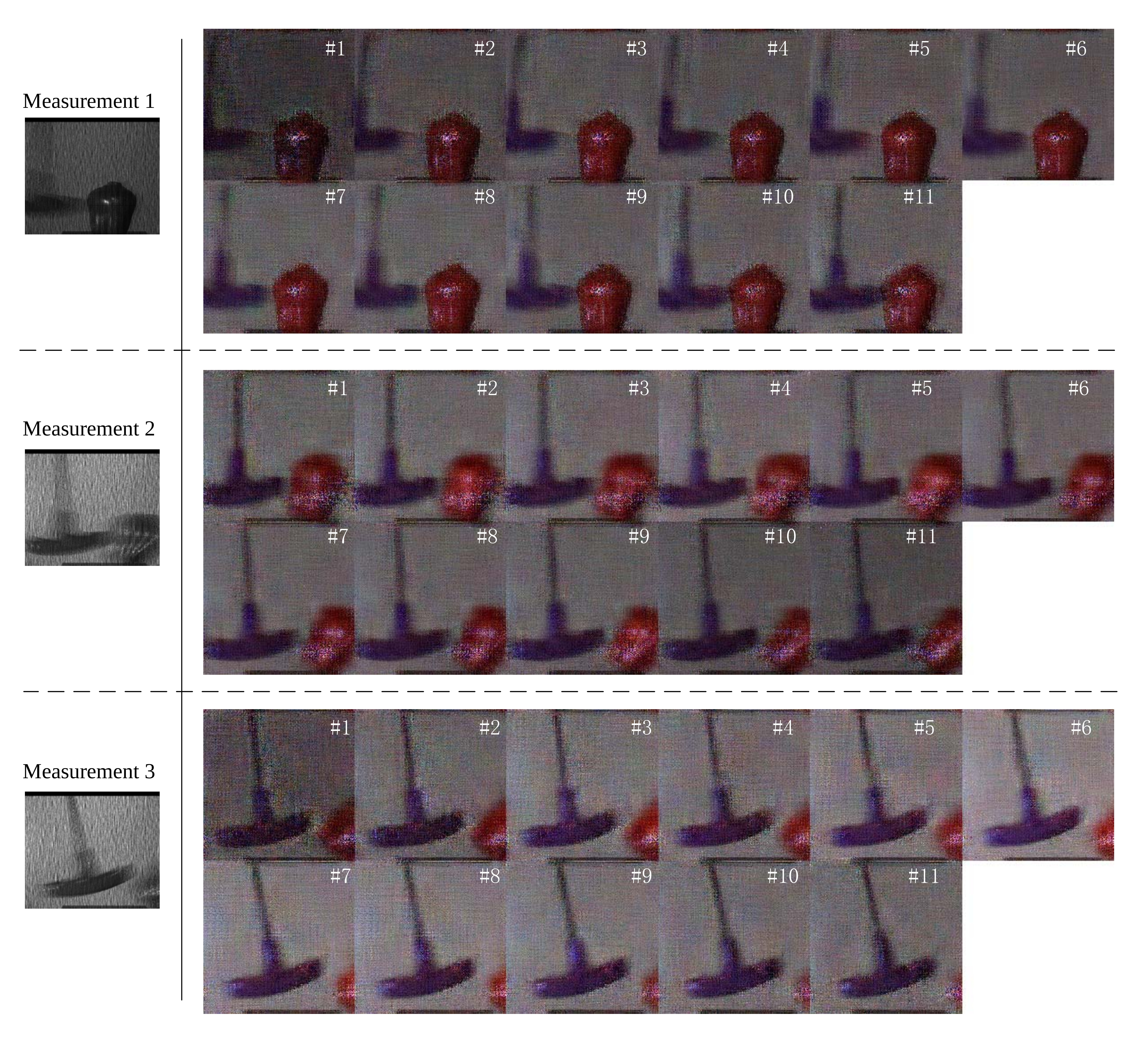}
\end{center}
\vspace{-6mm}
\caption{\small{Reconstruction results of a hammer hitting an apple.}}
\label{Fig:Hammer}
\vspace{-5mm}
\end{figure}

\vspace{-1mm}
We test our algorithm using real datasets captured by our color-CACTI camera, with selected results shown in Figures \ref{Fig:3Balls}-\ref{Fig:Hammer}.
Figure \ref{Fig:3Balls} shows low-framerate (captured at 30fps) compressive measurements of fruit falling/rebounding and corresponding high-framerate reconstructed video sequences. 
In the left are shown four contiguous measurements, and in the right are shown 22 frames reconstructed per measurement.
Note the spin of the red apple and the rebound of the orange in the reconstructed frames.
Figure \ref{Fig:Hammer} shows a process of a purple hammer hitting a red apple with 3 contiguous measurements.
We can see the clear hitting process from the reconstructed frames. 

\section{Conclusion}
We have implemented a color video CS camera, color-CACTI, capable of compressively capturing and reconstructing videos at low-and high-framerates, respectively.
A tree-structure Bayesian compressive sensing framework is developed for the video CS inversion by exploiting the 3D tree structure of the wavelet/DCT coefficients.
Both simulated and real datasets demonstrate the efficacy of the proposed model.

\clearpage
\newpage
\small
\bibliographystyle{IEEEbib}


\end{document}